\documentclass{article}

\usepackage{arxiv}

\usepackage[utf8]{inputenc} 
\usepackage[T1]{fontenc}    
\usepackage{hyperref}       
\usepackage{url}            
\usepackage{booktabs}       
\usepackage{amsfonts}       
\usepackage{nicefrac}       
\usepackage{microtype}      
\usepackage{xcolor}         
\usepackage{xspace}
\usepackage{graphicx}
\usepackage{subcaption}
\usepackage{float}
\usepackage{epsfig}
\usepackage{bbm}
\usepackage[ruled,vlined,linesnumbered,noresetcount]{algorithm2e}
\usepackage{algpseudocode}
\usepackage{amsmath}
\usepackage{xcolor}

\usepackage{booktabs}
\usepackage{multirow}
\usepackage{wrapfig}

\newenvironment{packeditemize}{
\begin{list}{$\bullet$}{
\setlength{\labelwidth}{8pt}
\setlength{\itemsep}{0pt}
\setlength{\leftmargin}{\labelwidth}
\addtolength{\leftmargin}{\labelsep}
\setlength{\parindent}{0pt}
\setlength{\listparindent}{\parindent}
\setlength{\parsep}{0pt}
\setlength{\topsep}{3pt}}}{\end{list}}

\DeclareMathOperator*{\argmax}{arg\,max}

\newcommand{\ours}{$\mathrm{D^2ULO}$\xspace}
\newcommand{\lsrc}{\mathcal{L}_s}
\newcommand{\utgt}{\mathcal{U}_t}

\newcommand{\uset}{S_{DS}}

\newcommand{\deepset}{DS}
\newcommand{\fewidx}{I}

\newcommand{\utilityFunc}{u}

\newcommand{\X}{\mathcal{X}}
\newcommand{\Y}{\mathcal{Y}}
\newcommand{\labeled}{\mathcal{L}}
\newcommand{\unlabel}{\mathcal{U}}
\newcommand{\targetSelectNum}{M}

\title{Zero-Round Active Learning}

\author{
    Si Chen \\
	Virginia Tech\\
	Blacksburg, VA \\
	\texttt{chensi@vt.edu} \\
	\And
	Tianhao Wang \\
	Harvard University\\
	Cambridge, MA \\
	\texttt{tianhaowang@fas.harvard.edu} \\
	
	\And
	Ruoxi Jia \\
	Virginia Tech\\
	Blacksburg, VA \\
	\texttt{ruoxijia@vt.edu} \\
}

\begin{document}
\maketitle
\begin{abstract}
Active learning (AL) aims at reducing labeling effort by identifying the most valuable unlabeled data points from a large pool. Traditional AL frameworks have two limitations: First, they perform data selection in a multi-round manner, which is time-consuming and impractical. Second, they usually assume that there are a small amount of labeled data points available \emph{in the same domain as} the data in the unlabeled pool. Recent work proposes a solution for one-round active learning based on data utility learning and optimization, which fixes the first issue but still requires the initially labeled data points in the same domain. In this paper, we propose \ours as a solution that solves both issues. Specifically, \ours leverages the idea of domain adaptation (DA) to train a data utility model which can effectively predict the utility for any given unlabeled data in the target domain once labeled. The trained data utility model can then be used to select high-utility data and at the same time, provide an estimate for the utility of the selected data. Our algorithm does not rely on any feedback from annotators in the target domain and hence, can be used to perform zero-round active learning or warm-start existing multi-round active learning strategies.
Our experiments show that \ours outperforms the existing state-of-the-art AL strategies equipped with domain adaptation over various domain shift settings (e.g., real-to-real data and synthetic-to-real data). Particularly, \ours is applicable to the scenario where the source and target labels have mismatches, which is not supported by the existing works. 
\end{abstract}

\section{Introduction}
Deep neural networks have been successful on various tasks across different fields with the help of large-scale labeled datasets. However, data labeling processes are often expensive and time-consuming. One popular framework to reduce labeling costs is \emph{active learning} (AL), which strategically selects and labels the data instances from the unlabeled data pool with the goal of achieving comparable performance with fewer labeled instances.


In a typical AL framework \cite{fine2002query, freund1997selective, graepel2000kernel, seung1992query, campbell2000query, schohn2000less, tong2001support,ash2019deep, sener2017active, wei2015submodularity, killamsetty2020glister, kirsch2019batchbald}, a learner begins with a small number of labeled data points and requests labels for more data points iteratively. At each round, a subset of points is selected based on its utility to the current model, which is trained on all the points selected in previous rounds. However, the multi-round nature could be a limitation for applying AL to real-world applications, because the most common labeling platforms, e.g., Amazon Mechanical Turk and annotation outsourcing companies, usually do not support a timely interaction between the learner and data annotators. Moreover, multi-round AL does not allow complete parallelization of labeling efforts, which can otherwise greatly improve annotation efficiency.

A recent work \cite{wang2021oneround} proposes a framework DULO that brings AL to a new setting where the selection is performed in only one round. Specifically, DULO starts by querying the labels for a small number of randomly selected unlabeled instances and after the annotator returns the labels for these instances are labeled, it trains a utility model that takes a set of points as input and outputs its corresponding utility. Utility of any possible set of unlabeled instances could be predicted by this model. Then, it selects the set of unlabeled instances by doing a greedy search to maximize the utility model. 
While the zero-round setting of DULO is attractive in practice, it still requires one-round interaction as it needs feedback from annotators to get the initial labeled data.
These initial labeled data are a basis for designing strategies for subsequent selection. 

In this paper, we explore the possibility of zero-round AL and ask the question: \emph{can we select unlabeled data in a way that does not rely on any feedback from potential annotators but works better than random selection?} Such data selection strategies, if exist, can be directly plugged into the widely used labeling platforms to help reduce labeling costs. Moreover, they can serve as a warm-start for existing AL approaches, one-round or multi-round, which all require a labeled data at the beginning. Our key idea to enable effective zero-round data selection is inspired by the observations that there are often labeled datasets available from related domains and in some applications, such as autonomous driving, there are off-the-shelf simulators that can simulate a large set of labeled data points that are related to the dataset to be labeled. Intuitively, these labeled data, although from a different domain, might still provide useful information about what types of data are worth being labeled.



In this paper, we present \underline{D}omain adaptive \underline{D}ata \underline{U}tility function \underline{L}earning and \underline{O}ptimization (\ours), an algorithm that can leverage labeled datasets from a source domain to help select the unlabeled data instances in the target domain. Importantly, our approach does not rely on any labeled instances from the target domain and hence provides a zero-round AL strategy. Specifically, we train a utility model that predicts the utility for any given unlabeled dataset, along with a feature extractor. We design our training scheme so that it will force the feature extractor to extract some domain-invariant features that are, at the same time, effective for predicting the utility of the dataset. One important benefit enabled by the utility modeling is that our approach can provide an estimate for utility of the selected data, which is very useful in practice for learners to decide the amount of unlabeled points to select and annotate. 
We apply \ours to unlabeled data selection on various object recognition tasks across domains. Experiments show that our algorithm achieves state-of-the-art results in various domain shifts settings, including real source/target-domain data as well as more challenging ones: source domain is synthetic data while target domain is real data, and source and target domain data have label mismatches.   

Compared with DULO, \ours has more novel applications. For example, as existing unsupervised domain adaptation often falls behind its supervised counterpart, it is necessary to select data points in the target domain to further improve the performance, and \ours provides a strategy that can select data more efficiently. Besides, our method can also be applied to perform data selection on the target domain even when it has inconsistent labels with the source domain, while typical AL strategies cannot. 

\section{Related Work}
\textbf{Active Learning.} 
\label{subsec: AL}
Active learning aims to reduce labeling effort by selecting data that are most valuable for model training, and it usually performs in an iterative manner. Earlier works \cite{fine2002query, freund1997selective, graepel2000kernel, seung1992query, campbell2000query, schohn2000less, tong2001support} select only one sample each round. Such AL strategies cannot parallelize labeling efforts and are often time-consuming in practice. Batch-mode active learning~\cite{ash2019deep}, by contrast, queries data in groups and hence improves learning efficiency; in particular, it can better handle models with slow training procedures (e.g., deep neural networks). Many other works investigated batch-mode AL as well. For example, Core-set \cite{sener2017active} performs k-center clustering to select informative data points while preserving their geometry. BADGE \cite{ash2019deep} attempts to capture the diversity and informativeness of data points in the gradient space and select data with gradients with diverse directions and high magnitude.
Some works \cite{wei2015submodularity, kirsch2019batchbald, killamsetty2020glister}, on the other hand, exploit the properties of submodular functions and hence further improve selection efficiency.


Different from AL strategies above which are designed to proceed iteratively until exceeding the labeling budget. Most recent work \cite{wang2021oneround} creates a new setting for active learning: they propose DULO for one-round AL, which selects the desired amount of unlabeled points all at once based on an initially labeled set. More specifically, DULO formulates the problem of one-round AL as the one of maximizing data utility functions, which map a dataset to some performance measure of the model trained on the set. In this paper, we propose \ours and show that it is possible to get rid of the use of labeled instances in the target domain by exploiting information from a related but different domain where annotated data is available. 


\textbf{Domain Adaptation.} 
Domain adaptation (DA) is a common solution to dealing with distribution shifts between source and target domain. The core idea is to learn some domain-invariant features, so the task model trained on the source domain can be readily applied to the target domain. There are three categories of DA depending on the data available from the target domain: unsupervised DA, semi-supervised DA, and supervised DA. Unsupervised DA is a setting where labeled target data is not available and agrees with the problem setting studied by our paper. Earlier works in this setting focus on minimizing some specific measurements of distributional discrepancy in the feature space. For example, \cite{long2015learning,long2017deep,tzeng2014deep} characterizes distribution distance via the Maximum Mean Discrepancy (MMD) of kernel embeddings; \cite{saito2018maximum,lee2019sliced} utilizes category predictions from two task classifiers to measure the domain discrepancy. These approaches were further improved by the use of an adversarial objective loss function regarding a domain discriminator that tries to distinguish between source and target feature embeddings \cite{tzeng2015simultaneous, ganin2015unsupervised, ganin2016domain, long2017conditional}.  However, the adversarial training may encounter the technical difficulty of model collapse \cite{mirza2014conditional}. A recent work \cite{hoffman2018cycada} combines generative adversarial networks (GAN) with cycle-consistent constraints and adapts representations at both feature-level and pixel-level effectively.


\textbf{Combining Active Learning and Domain Adaptation.}
Although both active learning and domain adaptation are two possible solutions to problem of insufficient labels, only a few works in the literature integrate these two methodologies into a single framework. \cite{chattopadhyay2013joint} proposes a method that re-weights source data and selects target data to query simultaneously, so that the dataset, consisting of re-weighted source samples, labeled target samples and queried target samples, is closest to the distribution of target unlabeled data. \cite{saha2011active,rai2010domain} propose ALDA that consists of three models: a domain adaptation classifier which adapts feature representation of source domain; a domain classifier that avoids querying labels for ``source'' data that are similar to target samples; a source classifier that provides labels for ``source`` data that resemble target samples. AADA, recently proposed by \cite{su2020active}, starts from training an unsupervised domain adaptation classifier, then target sample selection using importance weights. Model retraining is performed iteratively in AADA. These methods share the same limitation as most of the existing AL strategies as they are all designed to proceed for multi-rounds until exceeding the selection budget. To the best of our knowledge, \ours is the first that combines domain adaptation with AL in the zero-round setting.

\begin{figure}[t!]
\centering
\includegraphics[scale=0.3]{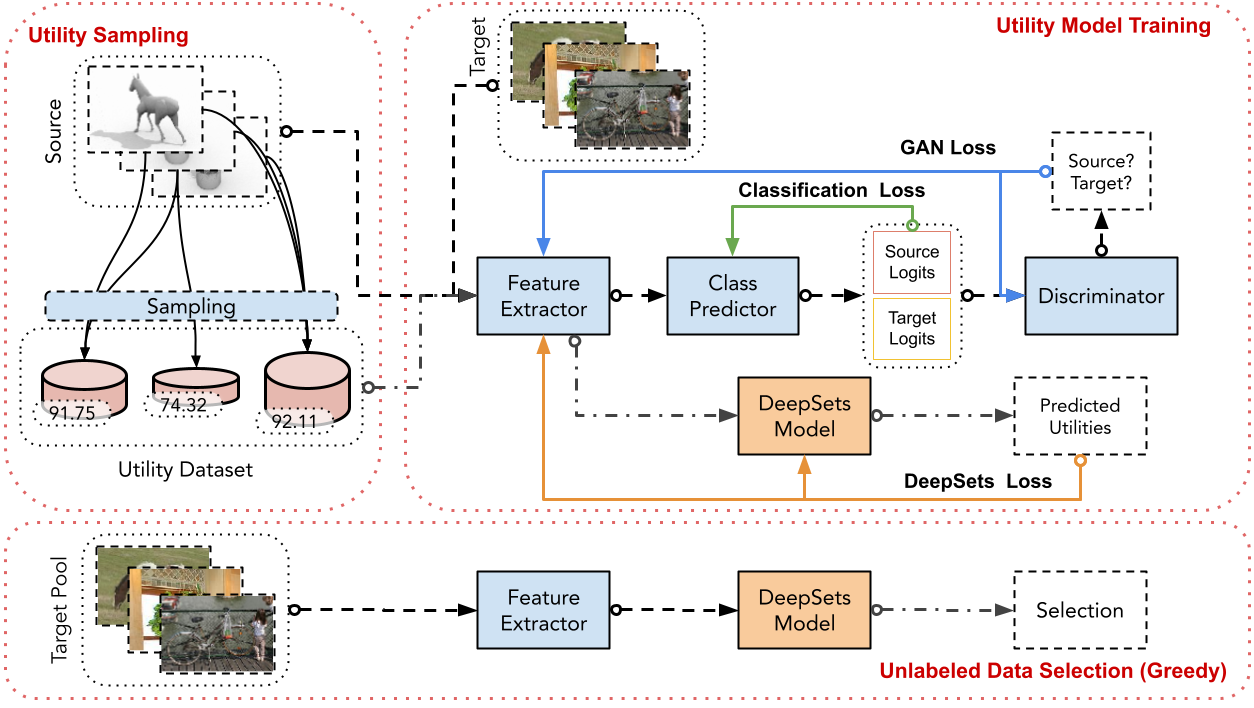}
\caption{Overall Workflow of \ours. 
}
\label{fig:workflow}
\end{figure}

\section{Approach}
\paragraph{Setting.} Existing AL strategies rely on a small amount of labeled data in the target domain $T$. By contrast, our goal is to develop zero-round AL strategies, which do not require any labeled data in the target domain.
We assume that there exists a source domain $S$ whose distribution $p_s$ is closely related to the target distribution $p_t$, while unlike target domain $T$, the label of instances from $S$ is already available or easily accessible.

In this section, we introduce our algorithm \ours. The key idea is to first learn a \emph{data utility model} that can predict the utility for any set of unlabeled instances. We leverage domain adaptation techniques to ensure that the model is useful for predicting the utility for unlabeled instances in the target domain and further use this model to guide the data selection. We will denote labeled data by $\labeled$, unlabeled data by $\unlabel$, and input and output spaces by $\X, \Y$, respectively.

\subsection{Overview}
\label{sec: u_learn}

\begin{algorithm}[t!]
\SetAlgoLined
\SetKwInOut{Input}{Input}
\SetKwInOut{Output}{Output}
\Input{a subset of samples $\labeled^I = (\X^I, \Y^I)$ chosen from training set which the index is given by $\fewidx$; validation set $\labeled^{val}$; classifier $f$; feature extractor $G_f$; metric function $\utilityFunc$.}
\Output{utility dataset $S_{\deepset}$ for DeepSets training.}

Initialize utility dataset ~$S_{\deepset} = \emptyset$.

\For{$i = 1, \dots, N$}{
    Randomly choose a subset $\labeled_i = (\X_i, \Y_i)$ where $\labeled_i \subseteq \labeled^I$.
    
    Train classifier $f$ with $\labeled_i$
    
    $u_i \leftarrow u(f, \labeled^{val})$
    
    $S_{\deepset} = S_{\deepset} \cup \{(\X_i, u_i) \}$.
}

\Return{$S_{\deepset}$}
 \caption{Data Utility Sampling}
 \label{alg:utilitylearning}
\end{algorithm}

The concept central to our AL strategy is a \emph{data utility function}, which maps any set of unlabeled instances to the performance of the ML model trained on the set once it is labeled. With such a function, AL can be done by simply selecting the unlabeled instances that maximize the output of the data utility model. Although data utility functions may have a close form for certain types of learning algorithms and model performance metrics (e.g., the test classification accuracy of K-Nearest-Neighbor \cite{wei2015submodularity}), for most models data utility functions cannot analytically derived. Recent work~\cite{wang2021oneround} proposed to learn data utility functions from data. Note that data utility functions are set functions, in which the input is a data set and the output is a real value indicating the utility of the data. Hence, each training samples for data utility function learning consist of a set of data points and the corresponding utility score, indicating the performance of the ML model trained on the set. Constructing the training set for data utility learning could be expensive, because to label each training sample, one needs to re-train the model. Fortunately, \cite{wang2021oneround} presents some empirical evidence that the learning of data utility functions could be sample-efficient due to its ``diminishing return'' property. Also, one can replace the original ML model with an efficiently-trainable proxy model (such as logistic regression) while still retaining good data selection performance. 

Note that data utility learning requires labeled data instances, which make it possible to create the training set. \cite{wang2021oneround} assumes a small labeled set in the target domain for data utility learning. However, this assumption no longer holds true in the zero-round AL setting. To resolve this problem, we propose to learn the data utility model on the source domain and mitigate the effects of domain shifts via domain adaptation. 


\subsection{\ours Algorithm}
The workflow of our algorithm is summarized in~\ref{fig:workflow}.

\paragraph{Utility Sampling.} The goal of this step is to construct the training set for learning data utility functions. Given a set of samples $\labeled^I$ and a validation set $\labeled_{val}$ in the source domain, each time we randomly sample a subset $\labeled_i \subseteq \labeled^I$ and train a classifier $f$ on it. Utility of this subset is then given by utility metric $u$ which in this paper is the validation accuracy of $f$ on $\labeled_{val}$. The utility training set $S_{\deepset}$ is thus $\{(\labeled_i, u_i)\}$. A general utility sampling workflow is demonstrated in Algorithm \ref{alg:utilitylearning}.

\paragraph{Utility model training.} The goal of this step is to train a utility model effective for predicting the utility for unlabeled data in the target domain. Following \cite{wang2021oneround}, we adopt the popular set function model--DeepSets~\cite{zaheer2017deep}--as the data utility model. DeepSets is a deep neural network which has the property of permutation invariance and equivariance, which make it suitable for set function modeling. Specifically, with the utility samples $\labeled$ from the last step, a feature extractor $G_f$ will be utilized to get the embedding of the training instances in $S_{\deepset}$, and the DeepSets model $f_{\deepset}$ maps the feature embedding of a set of points to its corresponding utility.




\begin{algorithm}[t!]
\SetAlgoLined
\SetKwInOut{Input}{Input}
\SetKwInOut{Model}{Model}
\Input{labeled source data $\lsrc = (\X_s, \Y_s)$; unlabeled target data $\utgt$; utility dataset $\uset=(X, U)$, where $X = (X_1, \ldots, X_N), U = (u_1, \ldots, u_N)$.}

\Model{$G=\{G_f, G_y, G_d\}$; feature extractor $G_f$; class predictor $G_y$; discriminator $G_d$; DeepSets utility model $f_{\deepset}$}


\For{$epoch = 1, \dots$}{
    \For{k steps}{
    Train $G$ with $(\lsrc,\utgt)$
    }
    
    Fix $G_f$; extract the feature embedding $E_s$ of utiltiy dataset  $E_s \leftarrow G_f(X)$
    
    Train a DeepSets model $f_{DS}$ on $(E_s, U)$
    
    Fix $f_{DS}$; train $G_f$ with $(X, U)$
    
}

\Return{$G_f$; $f_{DS}$}
 \caption{\ours}
 \label{alg:ZRAL}
\end{algorithm}

In the setting of interest to our paper, labeled data is not available in the target domain and the utility model $f_{\deepset}$ can only be trained on data from another domain. Hence, domain adaptation is needed to mitigate the performance drop caused by domain shift.

A domain adaptation framework usually consists of three components: a feature extractor $G_f$, a class predictor $G_y$ which takes the output embedding of $G_f$ and makes class predictions, and a discriminator $G_d$ that aims to distinguish between source and target domain data. DA typically has two goals: 1) map examples from two domains to a common feature space; and 2) retain useful information for classification. Those two goals are usually achieved through optimizing the GAN loss $L_{GAN}$ and classification loss $L_{cls}$, given by

\begin{align}
    L_{GAN}=-\big[\mathbb{E}_{x \sim p_s(x)} \log G_d(G_f(x))+ \mathbb{E}_{x \sim p_t(x)} \log (1-G_d(G_f(x)))\big]. 
\end{align} 

\begin{equation}
    L_{cls} = \text{CrossEntropy} \left(G_y(G_f(x)), y \right)
\end{equation}

We now discuss how to leverage domain adaptation in data utility learning to train a utility model useful for data selection in the target domain.
A naive way could be breaking the task into two steps: 1) apply domain adaptation techniques to obtain a feature extractor $G_f$ that extracts domain invariant features; and 2) train a DeepSets utility model $f_{\deepset}$ on the source feature extracted by the pre-trained $G_f$. However, the feature extractor learned in this way is only optimized towards the goal of being useful for classification, ignoring the goal of being useful for predicting data utility.
Hence, we propose a joint training process for $G_f$ and $f_{\deepset}$ that is mindful of the two goals simultaneously.

Specifically, given the labeled source data $\labeled_s$, unlabeled target data $\unlabel_t$, a utility training set $S_{\deepset}$ obtained from Algorithm \ref{alg:utilitylearning}, we alternate between $k$ steps of general domain adaptation training and one step of utility training. The former one just follows the usual DA framework. For the latter one, a DeepSets model $f_{\deepset}$ is first trained on $S_{\deepset}$ given current feature extractor $G_f$, and it will be fixed and used to optimize $G_f$ in turn given the same objective of minimizing DeepSets Loss:
\begin{align}
    \min_{f_{\deepset}}\min_{G_f} L_{DS} = \sum_{i=1}^N \Vert f_{\deepset}(G_f(X_i)) - u_i  \Vert^2
\end{align}

The main reason that we add $L_{DS}$ is that the features suitable for classification may not be equally helpful for learning data utilities. For example, the best possible features for classification tasks would be simply the label for the data points. However, this kind of features contains no information about the quality of the data points. Intuitively, $L_{DS}$ will enable $G_f$ to map source and target domain to a feature space that is more suitable for data utility learning.


Note that \ours can be combined with any state-of-the-art DA frameworks, and we use CyCADA \cite{hoffman2018cycada}, UDA \cite{sun2019unsupervised}, AFN\cite{Xu_2019_ICCV} in this paper.

\paragraph{Unlabeled Data Selection.} The last step of \ours is to seek for the unlabeled data attaining maximal utility under the learned utility model. Formally, we solve the following optimization problem:
\begin{equation}
\argmax_{|S|=\targetSelectNum, S \subseteq \unlabel} f_{\deepset} (G_f(S))
\label{eq:maximizingdeepset}
\end{equation} 
Similar to~\cite{wang2021oneround}, we perform a stochastic greedy algorithm to solve it.


\section{Evaluation}
\subsection{Evaluation Settings}
\subsubsection{Evaluation Protocol}
We use two approaches to evaluate the utility of the selected subset: 
\textbf{1) Train-from-Scratch:} we train a model from scratch on the data points selected from the target domain, and the utility of selected data points is given by the trained model's accuracy; 
\textbf{2) Fine-tune:} we adopt the fine-tune method proposed in \cite{zou2019consensus} to fine-tune the classifier $G_y$ obtained from our algorithm. Specifically, given a batch of labeled target samples chosen by the strategy, we compute the centroid of each class in the feature space and generate a hypothesized label for each unlabeled sample given its similarity between different centroids. We use the inverse of Wasserstein distance \cite{shen2018wasserstein} as the similarity metric.


\subsubsection{Baseline Algorithms}

For baseline algorithms, we combine state-of-the-art active learning strategies with domain adaptation. Specifically, we pre-train a feature extractor that minimizes the distance between source and target domain in the feature space, apply it to extract features for the unlabeled data pool and perform active data selection on the extracted features. Note that most of these existing AL strategies cannot be directly applicable to the zero-round AL setting. We limit the number of data selection of FASS, BADGE, GLISTER to be 1.

We compare \ours with the following state-of-the-art batch active learning strategies equipped with domain adaption. Specifically, our baselines contain as follows:
\begin{packeditemize}
    \item{\textbf{FASS.}} \cite{wei2015submodularity} performs subset selection as maximization of Nearest Neighbor submodular function on unlabeled data with hypothesized labels.
    \item{\textbf{BADGE.}} \cite{ash2019badge} selects a subset of samples with hypothesized labels whose gradients span a diverse set of directions.
    \item{\textbf{GLISTER.}} \cite{killamsetty2020glister} formulates the selection as a discrete bi-level optimization on samples with hypothesized labels.
    \item{\textbf{AADA.}} \cite{su2020active} uses a sample selection criterion which is the product of importance estimation and entropy of unlabeled data.
    \item{\textbf{Random.}} In this setting we randomly select a subset from all the unlabeled target data.
\end{packeditemize}
Moreover, we also train an ``optimal’' DeepSets model on labeled \emph{target} domain data, which serves as an upper bound of the active learning performance with only labeled source domain data available. We label this upper bound with \textbf{Optimal.} Note that this upper bound is \emph{not} realizable in the zero-round AL setting because of the lack of labeled target domain data. We plot this setting in the figures only to better understand how much room our strategy could be further improved.

\subsubsection{Datasets and Implementation Details}\label{sec:datasets}


\begin{table}[]
\centering
\begin{tabular}{ccccc}
\hline
\textbf{Source} & \textbf{Target} & \textbf{Domain Adaptation}  \\ \hline
MNIST\cite{lecun1998mnist}                            & USPS                            & CyCADA\cite{hoffman2018cycada}                                                     \\
USPS \cite{alpaydin1998optical}                            & MNIST                            & CyCADA                                                   \\
SVHN \cite{netzer2011reading}                            & MNIST                            & CyCADA                                                    \\
CIFAR-10 \cite{krizhevsky2009learning}                        & STL-10 \cite{coates2011analysis}                          & UDA \cite{sun2019unsupervised}           \\
VISDA-Synthetic \cite{visda2017}                        & VISDA-Real                       & AFN \cite{Xu_2019_ICCV}                                               \\
MNIST-04                          & MNIST-59                         & N/A                                            \\
MNIST-04                         & USPS-59                          & CyCADA                                          \\ \hline
\end{tabular}
\caption{Dataset and Training Settings. }
\label{tb:datasets}
\end{table}

Table \ref{tb:datasets} summarizes the datasets and implementation settings. 
We evaluate the performance of \ours and baseline approaches over four pairs of domain shifts: MNIST $\Rightarrow$ USPS, USPS $\Rightarrow$ MNIST, SVHN $\Rightarrow$ MNIST, CIFAR10 $\Rightarrow$ STL10. 
We also evaluate two more challenging transfer settings, where the source domain has inconsistent labels with the target domain: MNIST with digits 0-4 $\Rightarrow$ MNIST with digits 5-9, as well as MNIST with digits 0-4 $\Rightarrow$ USPS with digits 5-9. 
None of the baselines is applicable to these two settings by design. 
Following the settings in prior work~\cite{killamsetty2020glister,wang2021oneround}, we examine the effectiveness of different strategies on robust data selection, where partial data is corrupted by white noise. 


For all the source datasets, we randomly sample 300 (MNIST, USPS) or 500 (SVHN, CIFAR10, STL10) data points of the training set as $\labeled^I$ to perform data utility sampling demonstrated in Algorithm \ref{alg:utilitylearning}. We follow the implementation of DULO \cite{wang2021oneround} to set $N=5000$ and split the obtained $S_{\deepset}$ into training and validation set with a ratio $4:1$. We use small models (i.e., SVM, Logistic, Small CNN) as the classifier $f$ in Algorithm \ref{alg:utilitylearning} to obtain the data utility. This is because $f$ needs to be trained for thousands times to construct an utility dataset.
DULO \cite{wang2021oneround} empirically finds that data utility functions for small models are positively correlated with those for large models. Since data selection based on utility models only relies on the relative utility values between different set, utility models trained on samples obtained from small proxy models could still be useful for selecting data for large models.


We consider the state-of-art domain adaptation techniques for the specific transfer settings. Specifically, we combine our method with three different DA frameworks: CyCADA \cite{hoffman2018cycada}, UDA \cite{sun2019unsupervised}, and AFN\cite{Xu_2019_ICCV}, and the DA framework used for each transfer setting are given in Table \ref{tb:datasets}.
For training the DeepSets model, we use the same hyper-parameter as \cite{wang2021oneround}: we use Adam optimizer with learning rate $1e-5$,  mini-batch size of 32, $\beta_1$ = 0.9, and $\beta2$ = 0.999.

For Fine-tune performance evaluation, the starter model is $G_f$ in the corresponding domain adaptation framework in each setting. For Train-from-Scratch evaluation, we use three types of models to calculate the performance: 1) SVM, which is implemented with scikit-learn \cite{pedregosa2011scikit} with regularization parameter $C = 0.1$; 2) Logistic model; and 3)Small CNN model which has two convolutional layers and two max pooling layers and three fully-connected layers. Adam optimizer with learning rate $1e-3$, $\varepsilon=1e-7$ is used for training the small CNN model.

We use GeForce RTX 2080 ti for experiments on VISDA and NVIDIA Tesla K80 GPU for all the other experiments.

\subsection{Experiment Results}

\paragraph{Real-to-Real Adaptation.}
We start from comparing \ours with baselines on various domain shifts between real datasets. 

\begin{figure}[H]
    \centering
    \includegraphics[width=\textwidth]{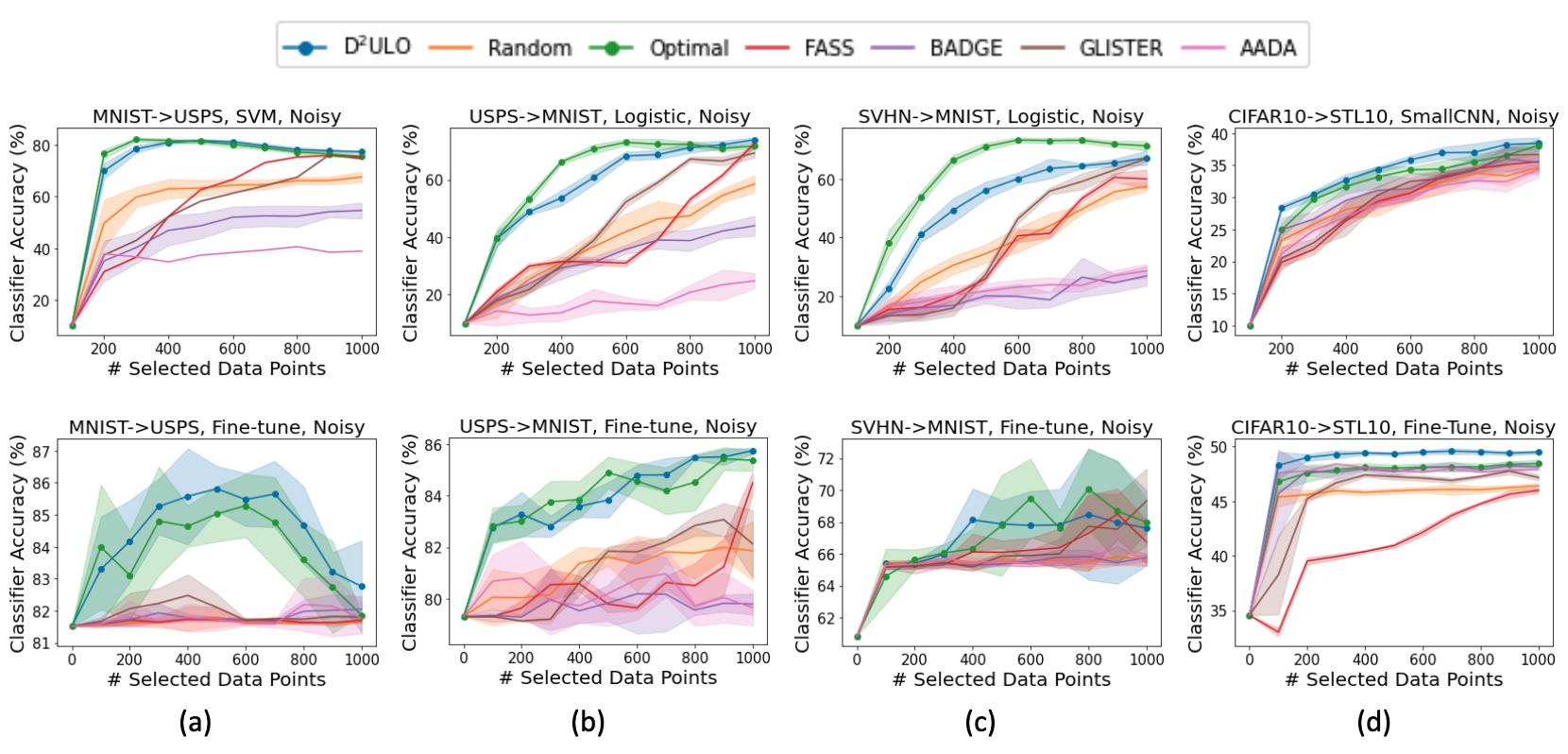}
    \caption{Performance of \ours on various adaptation shifts. The first row gives the results of Train-from-Scratch, where `SVM', `Logistic' and `SmallCNN' indicate the model used for obtaining utilities. The second row give the results of Fine-tune, and the start points are the classifier accuracy on target validation set after domain adaptation.}
    \label{fig:res-basic}
\end{figure}

Figure \ref{fig:res-basic} shows the results averaged over multiple random seeds. The $x$ axis shows the number of target sample selected by different strategies, and the $y$ axis is the accuracy of the model trained on selected points. 
In this figure, \ours outperforms all baselines in various shifts. Interestingly, the margin between \ours and Optimal is small, except for Figure \ref{fig:res-basic} (d) where Optimal is worse than \ours. This may be caused by the overfitting of DeepSets.


\begin{wrapfigure}{R}{0.3\linewidth}
    \centering
    \includegraphics[width=0.3\textwidth]{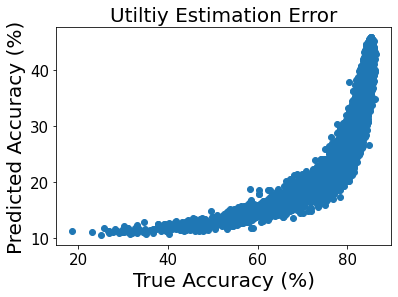}
    \caption{True Utiltiy vs Estimated Utility.}
    \label{fig:res-error}
\end{wrapfigure}

Another advantage of \ours over existing AL strategies is that we can provide a utility estimate for the selected data using the data utility model. Such a utility estimate could be very useful in practice for making an informed decision about the labeling budget. We compare our DeepSets estimated utility with the true utility for the MNIST $\Rightarrow$ USPS setting. Here, we randomly choose 300 data points of the target domain (USPS) and perform Algorithm \ref{alg:utilitylearning} to sample 4000 subsets. The true utilities are given by a SVM model trained on the subsets. Note that these 300 data points are unseen during DeepSets training. Figure \ref{fig:res-error} shows that the estimated utility is positively correlated with the true utility but systematically underestimates the true utility. Hence, the utility estimates provided by \ours can serve as a lower bound on the actual utility, which is still useful for guiding the choice of labeling budget. With better modeling of the relationship between the estimated and the true utility, one may be able to correct the bias in our estimation. We leave the exploration of this interesting direction to future work. Figure \ref{fig:res-error} also sheds light on the strong ability of \ours on differentiating unlabeled data quality in the target domain, even without access to any labeled data from the domain.


    


\paragraph{Synthetic-to-Real Adaptation.}
We further study the effectiveness of different strategies in the synthetic-to-real transfer setting. This setting could have great practical value because in many application domains, there exist sophisticated simulators that can generate a large amount of labeled data. We experiment on the VISDA-2017 dataset which has a significant synthetic-to-real domain gap. The source domain of VISDA are synthetic images generated by rendering from 3D models; the target domain are real object images collected from Microsoft COCO \cite{lin2014microsoft} and contains some natural variations in image quality.

\begin{figure}[t!]
    \centering
    \includegraphics[width=0.9\textwidth]{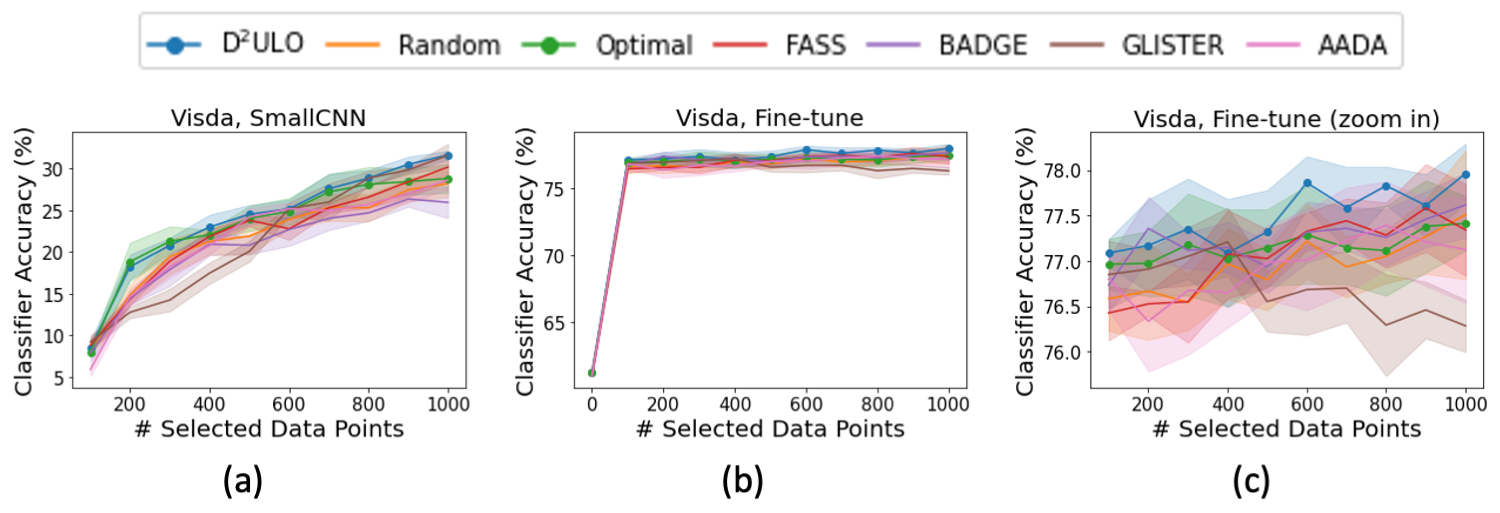}
    \caption{VISDA-2017 result: synthetic $\Rightarrow$ real. (a) gives the results of Train-from-Scratch, (b) and (c) give the result of Fine-tune. One interesting finding is that, all the strategies achieve a large performance improvement on classification accuracy. This indicates the needs to select data points from the target domain.}
    \label{fig:res-visda}
\end{figure}

As shown in Figure \ref{fig:res-visda}, \ours achieves the best performance among all the strategies. We also notice that even a very small amount of labeled target data can help improve the classifier accuracy by a large margin. For instance, in Figure \ref{fig:res-visda} (b), the Fine-tune accuracy increases rapidly at the beginning while only 100 data points are selected. This emphasizes the need of selecting data in the target domain for further improvement of domain adaptation performance. Since the target domain is relatively clean, random baseline works already very well. 

\begin{figure}[t!]
    \centering
    \includegraphics[width=0.6\textwidth]{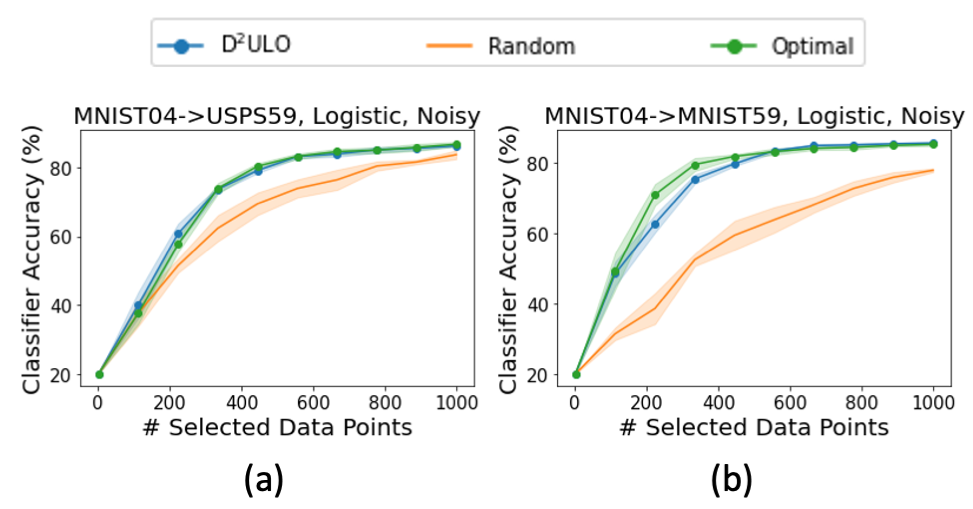}
    \caption{Performance of \ours on domains that have inconsistent label space.}
    \label{fig:res-label-trans}
\end{figure}

\paragraph{Label Mismatch.}
There are many real-world datasets that do not have overlap in the label space or only share a few common classes. Hence, we also conduct experiments in the setting where the source domain has entirely different object categories from the target domain. Specifically, we use digits 0-4 of MNIST dataset as the source domain, digits 5-9 of MNIST and USPS datasets as the target domain. This setting has great practical value yet has not been studied by previous AL literature. The reason is that most AL strategies rely on hypothesized labels generated by the classifier trained on the source domain and they become infeasible in this setting. For the same reason, we omit the Fine-tune performance metric and only report the accuracy of Train-from-Scratch. As we can see from Figure \ref{fig:res-label-trans},  both \ours outperforms random by a large margin and is comparable to Optimal.

\section{Conclusion}
\label{sec:conclusion}
In this paper, we propose \ours, a zero-round active learning strategy which does not require any labeled data in the target domain. 
We propose a novel training algorithm for data utility model, which extracts features from a data set useful for both classification and utility prediction.
We evaluate the effectiveness of \ours on various types of domain shifts and show that \ours can achieve state-of-the-art performance.

There are many interesting venues for future work. For instance, in our experiments, we observe that the DeepSet-based utility learning often overfits the training samples, which directly affects the efficacy of subsequent data selection tasks. One interesting future work is to develop preventative measures against overfitting for DeepSets training via a new training algorithm, model architectures, and regularization techniques. It is also interesting to study how to customize the synthetic data generation to the goal of improving active learning performance.


\bibliographystyle{unsrt}  
\bibliography{template}

\newpage

\appendix
\section{Details of Datasets Used in Section 4}
\paragraph{MNIST~\cite{lecun1998mnist}. }
MNIST dataset contains a training set of 60,000 examples and a test set of 10,000 examples. The images are grayscale handwritten digits with size $28 \times 28$. We resize the images to $32 \times 32$ in setting SVHN $\Rightarrow$ MNSIT.

\paragraph{USPS~\cite{alpaydin1998optical}. }
USPS dataset is a digit dataset scanned from envelopes. It contains a total of 9,298 $16 \times 16$ grayscale pixels. We resize them to $28 \times 28$ in both MNIST $\Rightarrow$ USPS and USPS $\Rightarrow$ MNIST setting.

\paragraph{SVHN~\cite{netzer2011reading}. }
SVHN is a real-world color house-number dataset containing 73,257 images for training and 26,032 images for testing. We use the version where all digits have been resized to $32 \times 32$ pixels.

\paragraph{CIFAR-10~\cite{krizhevsky2009learning}. }
The CIFAR-10 is an image recognition dataset containing 60,000 $32 \times 32$ 3-channel images in 10 classes. 

\paragraph{STL-10~\cite{coates2011analysis}. }
The STL-10 dataset consists of 13,000 color images of size $96 \times 96$ in 10 classes. We resize them to $32 \times 32$ in the experiments.

\begin{figure}[h!]
\centering
\includegraphics[scale=0.3]{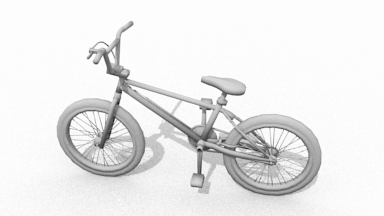}
\includegraphics[scale=0.2]{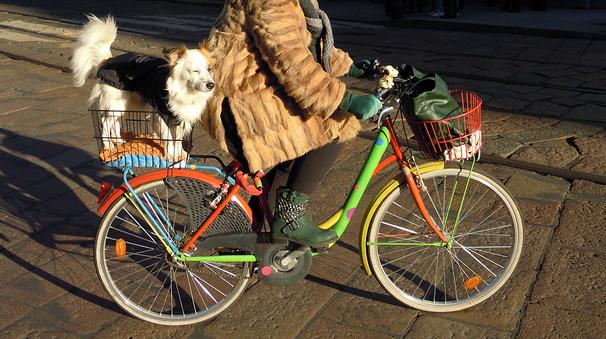}
\caption{Example images in VISDA2017. The left is an image of source domain (synthetic) while the right is an image of target domain (real).}
\label{fig:visda-example}
\end{figure}

\paragraph{VISDA2017~\cite{visda2017}. }
VISDA2017 dataset is designed for unsupervised domain adaptation challenge which contains more than 280K images across 12 object categories with large domain gaps. The source domain are synthetic 2D images rendering of 3D models which the angles and lighting conditions are different. The target domain are photo-realistic or real-images. In the experiment, we resize all the images to $256 \times 256$ and crop at the center obtaining images with size $224 \times 224$. An example of synthetic-real image pair is shown in Figure \ref{fig:visda-example}.

\section{Details of Models and Baseline Algorithms in Section 4}

\paragraph{SVM.}
We use Linear Support Vector Classification (SVC) implemented by scikit-learn \cite{pedregosa2011scikit} with L2 penalty and regularization parameter $C = 0.1$. Others remain as default.

\paragraph{Logistic Regression.}
We use Logistic Regression implemented by scikit-learn \cite{pedregosa2011scikit}. We set the maximum number of iterations to be 1000.

\paragraph{Small CNN.}
The small CNN model we used has two convolutional layers and two max pooling layers and three fully-connected layers. We use Adam optimizer with learning rate $10^{-3}$, $\varepsilon=10^{-7}$, batch size 32 for training the small CNN model.

\paragraph{DeepSets Model.}
A DeepSets model can be represented as $f_{\deepset}(S) = \rho ( \sum_{x \in S} \phi(x))$ where both $\rho$ and $\phi$ are neural networks. In our experiments, both $\rho$ and $\phi$ contain 3 linear layer with ELU activation, and we set the number of neurons to be 256 in each hidden layer, the dimension of set features which is the output of $\phi$ network to be 256. For training DeepSets models, we use Adam optimizer with learning rate $10^{-5}$, batch size 32, $\beta_1=0.9$, and $\beta_2=0.99$. 

\paragraph{Baseline AL Techniques.}
We use BADGE, FASS, and GLISTER implemented by DISTIL\footnote{\url{https://github.com/decile-team/distil}}. Specifically, we set batch size to be 32 for all of the three strategies, and learning rate to be 0.001 for glister.

\section{Other Implementation Details}
\paragraph{Domain Adaptation.}
We test our method with three state-of-the-art domain adaptation frameworks in this paper: CyCADA \cite{hoffman2018cycada}, UDA \cite{sun2019unsupervised}, AFN \cite{Xu_2019_ICCV}. 

For CyCADA\footnote{\url{https://github.com/jhoffman/cycada_release}}, we follow their official implementation where a source classifier is firstly trained using Adam optimizer with learning rate $10^{-4}$, batch size 128, $\beta_1=0.9$, and $\beta_2=0.99$. Then, weights of this source classifier are used as the initial weights of target classifier to perform domain adaptation. The same optimizer is used for training target classifier. We set the $k$ in Line 10 of Algorithm 2 to be 10.

For UDA\footnote{\url{https://github.com/yueatsprograms/uda_release}}, we use SGD optimizer with initial learning rate 0.1. We later decay the learning rate to 0.001 after 10 epochs. And we set $k$ to be 5.

For AFN\footnote{\url{https://github.com/jihanyang/AFN}}, we use SGD optimizer with learning rate 0.001 and weight decay $5\times 10^{-4}$ for training feature extractor, and SGD optimizer with learning rate 0.001, momentum 0.9 and weight decay $5\times 10^{-4}$ for training class predictor. We set $k$ to be 5.

When integrating all of the above three DA frameworks into D$^2$ULO, we use the same Adam optimizer with learning rate $10^{-6}$, $\beta_1=0.9$, and $\beta_2=0.99$ for DeepSets Loss back-propagation. 

\paragraph{Data Selection.}
We apply stochastic greedy optimization \cite{mirzasoleiman2015lazier} to solve Equation (4), and we set $\epsilon=10^{-3}$.

\end{document}